# Tongue image constitution recognition based on Complexity Perception method

Jiajiong Ma[a], Guihua Wen [a], Yang Hu[a], Tianyuan Chang[a], Haibin Zeng[a], Lijun Jiang[a], Jianzeng Qin[b,*]

Correspondence should be addressed to Jianzeng Qin[c,*]

[a]School of Computer Science and Engineering, South China University of Technology, Guangzhou 510000, China

[b]Southern Medical University, Guangzhou 510000, China

## Abstract

Background and Object：In China, body constitution is highly related to physiological and pathological functions of human body and determines the tendency of the disease, which is of great importance for treatment in clinical medicine. Tongue diagnosis, as a key part of Traditional Chinese Medicine inspection, is an important way to recognize the type of constitution. In order to deploy tongue image constitution recognition system on non-invasive mobile device to achieve fast, efficient and accurate constitution recognition, an efficient method is required to deal with the challenge of this kind of complex environment.

Methods：In this work, we perform the tongue area detection, tongue area calibration and constitution classification using methods which are based on deep convolutional neural network. Subject to the variation of inconstant environmental condition, the distribution of the picture is uneven, which has a bad effect on classification performance. To solve this problem, we propose a method based on the complexity of individual instances to divide dataset into two subsets and classify them separately, which is capable of improving classification accuracy. To evaluate the performance of our proposed method, we conduct experiments on three sizes of tongue datasets, in which deep convolutional neural network method and traditional digital image analysis method are respectively applied to extract features for tongue images. The proposed method is combined with the base classifier Softmax, SVM, and DecisionTree respectively.

Results：As the experiments results shown, our proposed method improves the classification accuracy by 1.135% on average and achieves 59.99% constitution classification accuracy.

Conclusions：Experimental results on three datasets show that our proposed method can effectively improve the classification accuracy of tongue constitution recognition.

*Keywords*：*Tongue diagnosis; Constitution classification; Complexity; Deep* learning

## 1. Introduction

China has a vast and increasing population. Subject to the shortage of medical resources, how to relieve medical pressure and make full use of medical resources is especially important in China (Yang et al., 2016). With the development of mobile technology and communication technology, mobile health (mHealth) technology enables patients to communicate with doctors at a long distance and obtain high-quality medical services. It is convenient for monitoring and managing patients (Li et al., 2014; Ni et al., 2014), and even for using mobile devices to diagnose disease (Huang et al., 2014).  In China, Body constitution is described as a form of innate and acquired talent in the process of human life. It is a comprehensive expression of physiological function and psychological state (Wang et al., 2011). The type of constitution is highly related to some diseases and even determines the tendency of the disease (Liu et al., 2016; Wang et al., 2013; Yu et al., 2016), which is of great importance for treatment in clinical medicine. Tongue diagnosis is an important way to recognize the type of constitution (Peng et al., 2017). Recognizing the patients' constitution through tongue images provide a simple, fast and cheap treatment of disease.

Since "Classification and Determination of Constitution in TCM" (China Association of Chinese Medicine, 2009) was officially published in 2009, constitution types were usually determined by Traditional Chinese Medicine constitutional questionnaire that may confuse patients. Beyond that, some studies utilized biomedical and computer technologies to achieve more accurately and quickly determination of constitution type ( Yang et al., 2011; Zhang et al., 2016). However, to the best of our knowledge, there is no research on automatic constitution classification by tongue images and most of works focus on statistically significant correlation between the tongue image and the constitution type. Tongue diagnosis, as an important and unique part of Traditional Chinese Medicine inspection, can examine the physiological and pathological changes of human body by observing the color and texture of tongue (Han et al., 2014; Lo et al., 2015). Besides that, tongue diagnosis has been used to recognize constitution type for many years. In this paper, we aim to use tongue images to automatically classify patients into nine constitution types: Qi-deficiency, Yin-deficiency, Yang-deficiency, Phlegm-wetness, Wetness-heat, Qi-depression, Blood-stasis, Special diathesis and Gentleness (China Association of Chinese Medicine, 2009). This work is divided into four steps: tongue image acquisition, tongue main area detection & calibration, tongue feature extraction and constitution classification. Previous works mainly used digital tongue imaging system (Kim et al., 2009; Zhang et al., 2005) or hyperspectral tongue imager (Li et al., 2010; Liu et al., 2007; Zhi et al., 2007) to capture tongue images. However, these professional devices are expensive, and lack robustness to the requirements of camera (lighting, shooting angle, resolution, etc.), and do not integrate cumbersome post-processing into acquisition device, which cannot take full advantage of mHealth. Previous works focused on tongue body area segmentation (Ning et al., 2012; H. Zhang et al., 2006) and did pattern recognition by extracting hand-crafted color, texture and geometry features (Juyeon Kim et al., 2014; Wang et al., 2013). However, only segmenting tongue image may lose boundary information which may be important for tongue diagnosis. Compared to deep learning methods, traditional image classification methods cannot effectively improve the accuracy with hand-crafted image features. Because of the excellent performance of deep learning in biomedical images detection (Su et al., 2017; Zhou et al., 2017; Zia ur Rehman et al., 2018) and classification (Anthimopoulos et al., 2016; Keivanfard et al., 2010; Sharma et al., 2017), we utilize Faster R-CNN (Ren et al., 2017) to detect tongue area and use calibration network inspired by CNN-based face bounding box calibration (Li et al., 2015) to further adjust detection area. Subsequently, we perform the constitution classification using deep convolutional neural network methods to extract features. In this work, we deploy tongue image constitution recognition system on non-invasive mobile device, which can capture patients' tongue images and upload them to the cloud server that achieves tongue area detection, tongue area calibration and constitution classification. After that, mobile device will receive the result of constitution type. In this way, we can provide convenient and quick constitution judgment on tongue diagnosis system. The flow diagram of tongue image constitution recognition is shown in Fig.1. The details are discussed in next section.

Unlike previous works that only worked in a standard environment, a challenge in our work is that capturing tongue images by mobile devices may face problems with various environmental condition (lighting, shooting angle, resolution, etc. ), which will lead to unevenness of the images distribution and the disparity between the complexities of the image features. A common method for machine learning is to fit a model by all training data and use it for constitution recognition. In this way, images with disparate complexity are used together to train the model, which have a bad effect on model training (Garcia et al., 2015) and performance of constitution recognition. Several studies had focused on characterizing the complexity of the entire dataset. However, they did not consider the complexity of individual instances (Smith et al., 2014) and provide solutions. In order to solve this problem, we propose Complexity Perception (CP) method that separate dataset into two disjoint subsets called easy data and difficult data according to their learning complexity at the individual level, and then train two classifiers separately. In this way, data with disparity complexity does not interfere with each other. For a testing sample, we distinguish its complexity and classify it by the corresponding classifier. In order to simulate the real scene where our tongue diagnosis system is used, we employ 50 volunteers to capture patients' tongue images in Traditional Chinese Medicine Department in Hospital of Guangdong Province by using smart phone. We construct three size of tongue images datasets and do experiment on these datasets to evaluate the effectiveness of the proposed method. As experimental results shown, our proposed method can effectively improve the accuracy of the classification of constitution.

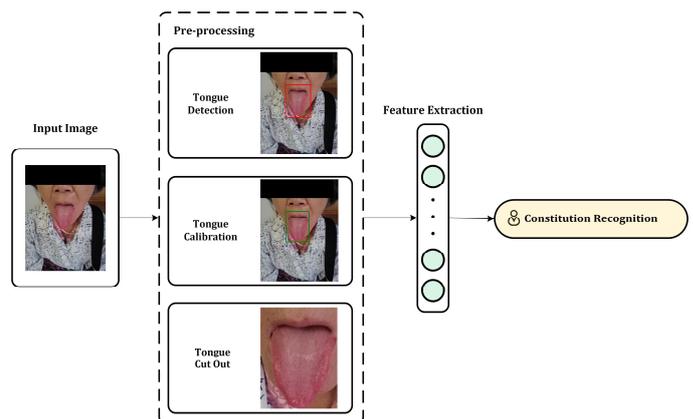

**Fig 1** A flow diagram of tongue image constitution recognition diagnosis system

This work has two major contributions. First, it utilizes a framework to deploy tongue image constitution recognition system, including tongue image acquisition, tongue main area detection & calibration, tongue feature extraction and constitution classification on non-invasive mobile device to reduce the pressure of the allocation of medical resources. Secondly, it proposes a new method to consider the complexity of instance at individual level, so as to reduce the effects of the uneven distribution of images caused by various environmental conditions, e.g., lighting and resolution. The proposed Complexity Perception method can be easily extended to other datasets and application scenarios, which has strong research significance.

## 2. Method

In this section, the design of tongue image constitution recognition are described. Part 1 briefly introduces how the images are acquired by tongue image acquisition system. In order to obtain high-quality tongue images and avoid the interference of background, which is of great importance for accurate tongue feature extraction and constitution recognition, the tongue detection and calibration pre-process are presented in Part 2. Part 3 describes the feature extraction frameworks. In Part 4, we discuss Complexity Perception method in detail.

*2.1 Tongue image acquisition*

We deploy a tongue images acquisition system on mobile device and employ 50 volunteers to capture patients' tongue images in Traditional Chinese Medicine Department in Hospital of Guangdong Province. Each image is annotated by an experienced expert on the same standard. Up to now, we have acquired 22482 tongue images and parts of them are used to train detection and calibration model. Finally, all these images are used to build a tongue images constitution recognition database.

*2.2 Tongue detection & calibration*

As the Fig.1 shown, in order to exclude the influence of extraneous part and better extract the features of the tongue, we need to extract the tongue area from the whole image. In this work, an object detection algorithm based on deep convolutional neural network is applied to detect the tongue area from image.

Subject to the high time complexity, windows redundancy and hand-craft features, deep convolutional neural network is widely used in object detection instead of traditional methods. Faster R-CNN applies the region proposal network (RPN) to replace Selective Search, so that it shares convolutional features with the detection network and enables nearly cost-free for the calculation of region proposals. Therefore, Faster R-CNN is applied to achieve tongue detection. We use the first 13th layers of VGG-16 (Simonyan and Zisserman, 2014) model to construct detection network and the model parameters are initialized with the pre-train weights on the ImageNet. RPN applies sliding window to map the output of convolutional features to a 512-d vector. And then this vector is fed into box-classification layer (cls) and box-regression layer (reg), which are fully-connected layers. Cls layer is trained to distinguish foreground and background, while reg layer can predict region proposal for each anchor. We use 4 scales and 3 aspect ratios to predict $k=12$ anchors at each sliding-window location.

After using tongue detection based on Faster R-CNN, we utilize calibration network that determines the category of the offset of detection area and calibrates the detection area to improve the result of tongue detection. We adjust the detection window $(x, y, w, h)$ with top-left corner at $(x, y)$ of size $(w, h)$ according to Equation (1), (2).

$$(x - \frac{x_n w}{s_n}, y - \frac{y_n h}{s_n}, \frac{w}{s_n}, \frac{h}{s_n}) \quad (1)$$

$$x_n \in \{-0.17, 0, 0.17\}$$
$$y_n \in \{-0.17, 0, 0.17\} \quad (2)$$
$$s_n \in \{0.83, 0.91, 1.0, 1.10, 1.21\}$$

*2.3 Feature extraction framework*

*2.3.1 Deep Neural network*

In recent years, methods based on deep convolutional neural networks have become widely used in the field of segmentation, detection and classification in computer vision. Therefore, some methods based on deep convolutional neural networks are applied to extract features in this paper. These methods have done well in ILSVRC (ImageNet Large Scale Visual Recognition Competition) that evaluates algorithms for images classification and object detection such as AlexNet (Krizhevsky et al., 2012), VGG (Simonyan and Zisserman, 2014), GoogLeNet (Szegedy et al., 2015), ResNet (He et al., 2016), and SENet (Hu et al., 2017).

In our case, we modify the last few layers in ResNet-50, VGG-16 and Inception-V3 model, and then train the them on tongue images dataset. Convolutional neural network can be regarded as a complex feature extraction function that can effectively extract important features in the tongue images. Therefore, the output of the second-to-last layer of neural network $x_i$ is used to represent the tongue image $X_i$ according to Equation (3) and (4).

$$x_{i,conv} = f_{conv}(X_i, W_{conv}) \quad (3)$$

$$x_i = f_{mlp}(x_{i,conv}, W_{mlp}) \quad (4)$$

Where, $f_{conv}$ is several convolutional layers, $W_{conv}$ is the weight of these convolutional layers, $f_{mlp}$ is two fully-connected layers and $W_{mlp}$ is the weight of fully-connected layers.

*2.3.2 LBP & Color moment*

In addition to deep learning methods, traditional digital image processing methods can also be used to extract the features of tongue images. The changes in the texture and color of the tongue tend to reflect changes of the state of the body. That is why we combine texture and color features to represent tongue images. The Local Binary Pattern (LBP) (Ojala et al., 1996) algorithm, which has been widely used in various application, is applied to describe the local texture features of tongue images. We split tongue image into $L \times L$ regions and utilize LBP on each region with $P$ sampling points on $R$ radius circle according to Equation (5) and (6). Here, $g_c$ is the value of center pixel and $g_p$ is the value of sampling points.

$$LBP_{P,R} = \sum_{p}^{P-1} s(g_p - g_c) 2^p \quad (5)$$

$$s(z) = \begin{cases} 1, & z \geq 0 \\ 0, & z < 0 \end{cases} \quad (6)$$

We compute the histogram of LBP features and concatenate them. And then the dimensions of concatenated features are reduced using Principal Component Analysis (PCA), following Equation (7). $ht_l$ is the LBP histogram of the $l^{th}$ region.

$$P_L(X_i) = PCA([ht_1, ht_2, \cdots, ht_{L^2}]) \quad (7)$$

The first three Color-Moments (Stricker and Orengo, 1995) in terms of mean, variance and skewness of each color channel are applied to represent the distribution of tongue image according to Equation (8), (9), (10) and (11). Here, the value of the $u^{th}$ color channel at the $v^{th}$ image pixel is $p_{uv}$.

$$E_u = \frac{1}{N} \sum_{v=1}^{N} p_{uv} \quad (8)$$

$$\sigma_u = (\frac{1}{N} \sum_{v=1}^{N} (p_{uv} - E_u)^2)^{\frac{1}{2}} \quad (9)$$

$$s_u = (\frac{1}{N} \sum_{v=1}^{N} (p_{uv} - E_u)^3)^{\frac{1}{3}} \quad (10)$$

$$P_C(X_i) = [E_1, E_2, E_3, \sigma_1, \sigma_2, \sigma_3, s_1, s_2, s_3] \quad (11)$$

Finally, the feature of tongue image $X_i$ is represented as the combination of texture and color features, following Equation (12).

$$x_i = [P_L(X_i), P_C(X_i)] \quad (12)$$

### 2.4 Complexity Perception Method

The detailed process of the Complexity Perception method will be described in three steps. A schematic diagram of Complexity Perception method is shown in Fig.2.

The first step, training data will be separated into two disjointed datasets called easy training data and difficult training data according to their learning complexity at the individual level. We need to train several classifiers on training data and calculate the accuracy of each training sample in these classifiers. According to the accuracy of each training sample in these classifiers, high-accuracy samples are divided into easy training dataset while the low-accuracy samples are divided into difficult training dataset. In our implementation, we randomly separate training data into k-fold and use constitution type label to train the classifiers (e.g. softmax, svm) on a fold dataset each time, which can obtain $k$ classifiers. The whole process above can be repeated $e$ times and obtain $N = e \times k$ classifiers. And then we separate training data into two datasets according to Equation (13), (14)

$$E_{train} = \{(x_i, +) \mid \frac{N(x_i)}{N} \geq \theta\} \quad (13)$$

$$D_{train} = \{(x_i, -) \mid \frac{N(x_i)}{N} < \theta\} \quad (14)$$

Here, $E_{train}$ is the easy training dataset and $D_{train}$ is the difficult training dataset. $N(x_i)$ indicates the number of times that a sample $x_i$ is predicted correctly among $N$ classified times. The threshold $\theta$ is a hyper-parameter that indicates the easy degree of training sample. According to Equation (13) and (14), if the accuracy of a training sample among $N$ classified times is greater or equal to $\theta$, it can be regarded as an easy training sample with low complexity, otherwise it is a difficult training sample with high complexity.

Next step, we apply machine learning method to distinguish the complexity of testing samples. It needs a new training database containing two class labels, which is defined as follows, where and $C_i$ is the complexity label of the sample $x_i$.

$$T_{train} = E_{train} \cup D_{train} = \{(x_i, C_i) \mid C_i \in \{+, -\}\} \quad (15)$$

$$T_j = \{(x_1, C_1), (x_2, C_2), \cdots, (x_n, C_n)\} \subseteq T_{train} \quad (16)$$

$$h_\beta(x) = \phi(\beta^T x) = \frac{1}{1 + e^{-\beta^T x}} \quad (17)$$

$$C_j = \begin{cases} +, & h_\beta(x) < 0.5 \\ -, & h_\beta(x) > 0.5 \end{cases} \quad (18)$$

Based on the training database $T_{train}$, we combine a hybrid model composed of K-Nearest Neighbor (KNN) and Logistic Regression (Logit) to construct complexity discriminator. For each testing sample $x_j$, we utilize KNN to find $n$ nearest neighbors $T_j$ in training dataset $T_{train}$ and use this dataset $T_j$ with complexity label to train a local Logistic Regression classifier called as $Clf_{C,j}$. This local classifier is utilized to determine the complexity of testing sample and classify it into easy testing sample or difficult testing sample with Equation (17) and (18), where $\beta^T$ is the parameter of Logistic Regression classifier $Clf_{C,j}$.

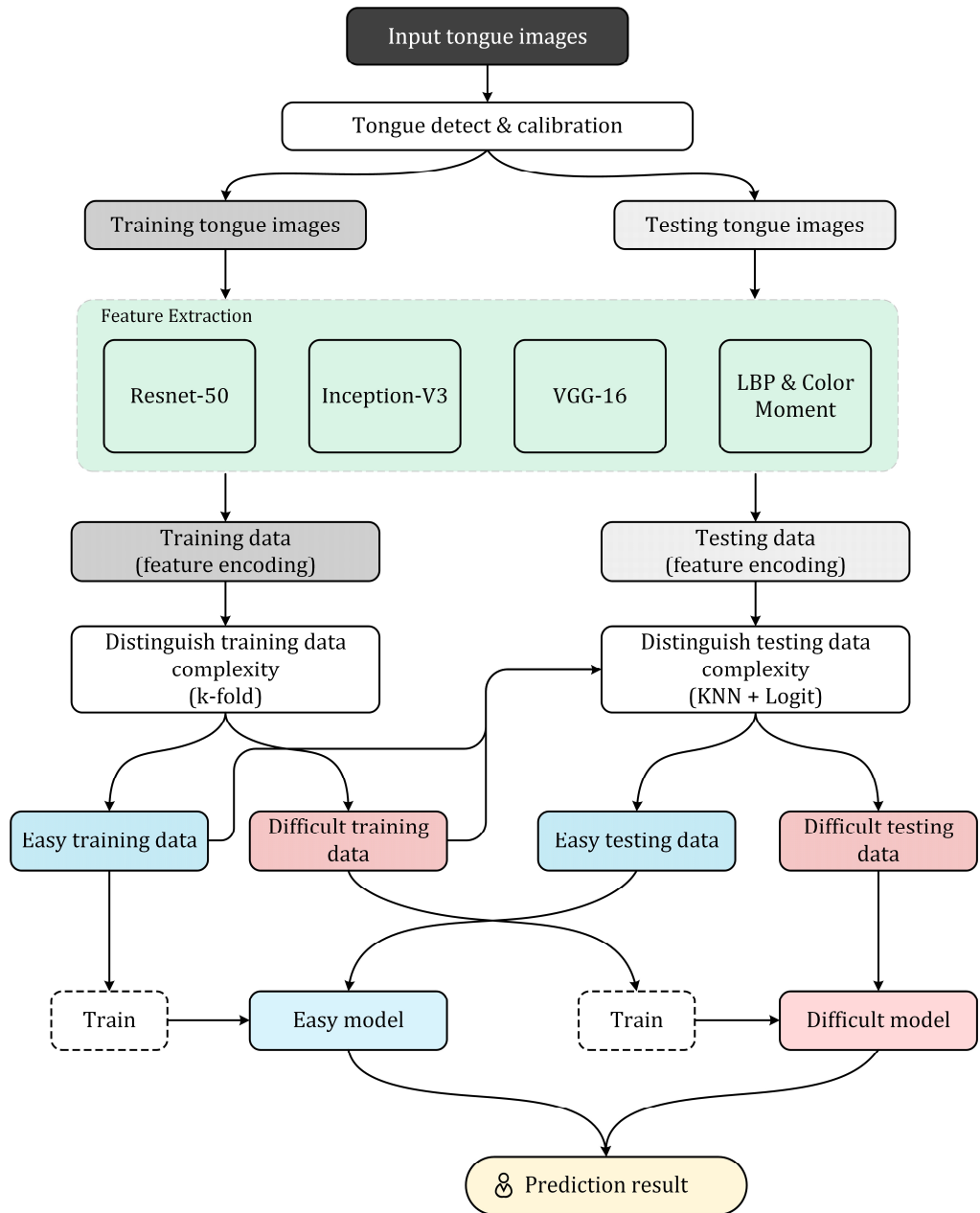

**Fig.2** Complexity Perception method

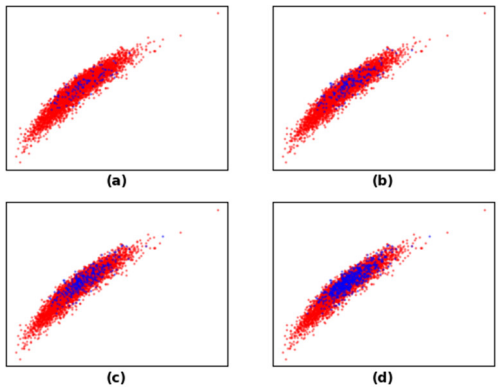

**Fig. 3** The testing samples distribution (2-dim). The red and blue points represent easy samples and difficult samples. Threshold $\theta$ for (a) ~ (d) is 0.6, 0.7, 0.8 and 0.9, respectively.

As a result of different environment (i.e. lighting), the features of different samples extracted by same method may quite different. We think that the features of easy samples represent tongue images well, while the features of difficult samples do not show the important information of tongue. In other words, easy samples have more significant distribution difference than difficult samples. As shown in Fig.3, there is a more dense distribution of difficult samples (blue points) than easy samples in the feature space. The problem of excessively dense distribution of difficult samples will lead to the hardness of classification. Training model on the whole data with disparate complexity will have a bad effect on the performance of model. Therefore, in the last step, we train easy classifier $Clf_E$ on easy training data and difficult classifier $Clf_D$ on difficult training data. In this way, data with disparity complexity does not interfere with each other. Finally, $Clf_E$ is used to predict easy testing samples and $Clf_D$ is used to predict difficult testing samples. The whole method is described as Algorithm 1.

**Algorithm 1** Complexity Perception Method for tongue image constitution recognition

---

**Require:** training data $(x_i, y_i)$, testing sample $x_j$, easy degree threshold $\theta$, and other parameters $k$, $e$ and $n$

**outputs:** the class label $\omega$ for the testing sample $x_j$

**Begin:**
1:  $N(x_i) = 0$ for all $x_i$, $N = e \times k$
2:  **for** iteration from 1 to $e$ **do**
3:   separate training data into $k$ Fold
4:   **for** subsamples from $k$ Fold **do**
5:    train classifier $Clf: x_i \to y_i$ on the subsamples
6:    **for** each sample $x_i$ in training data **do**
7:     $\varepsilon \leftarrow Clf(x_i)$
8:     if $y_i = \varepsilon$, $N(x_i) += 1$
9:    **end for**
10:   **end for**
11: **end for**
12: determine easy training samples
$$E_{train} = \{(x_i, +) \mid \frac{N(x_i)}{N} \geq \theta\}$$
13: determine difficult training samples
$$D_{train} = \{(x_i, -) \mid \frac{N(x_i)}{N} < \theta\}$$
14: train classifier $Clf_E : x_i \to y_i$ for $x_i \in$ easy training samples
15: train classifier $Clf_D : x_i \to y_i$ for $x_i \in$ difficult training samples
16: find $n$ nearest neighbors for $x_j$ in training data space $T_{train}$
$$T_j = \{(x_1, C_1), (x_2, C_2), \cdots, (x_n, C_n)\}$$
17: train local classifier
$$Clf_{C,j} : x_i \to C_i, (x_i, C_i) \in T_j$$
18: $C_j \leftarrow Clf_{C,j}(x_j)$
19: if $C_j \in \{+\}$, $\omega \leftarrow Clf_E(x_j)$
20: if $C_j \in \{-\}$, $\omega \leftarrow Clf_D(x_j)$

## 3. Results

### 3.1 Dataset Description

After more than a year of acquiring images in hospitals, we have collected 22482 tongue images. All these images are processed according to the second section, including tongue area detection and tongue area calibration, so as to exclude the influence of extraneous part of images. The distribution of categories of constitution types is shown in Table 1.

In order to observe the effect of Complexity Perception method under different sizes of dataset, we randomly select 100%, 80% and 60% tongue images to build three datasets, respectively denoted as Tongue-100, Tongue-80 and Tongue-60. Due to the imbalance of samples, the random selection is carried out in each category.

**Table 1** The distribution of categories of constitution types

| Class Name | #Samples | |
|---|---|---|
| Qi-deficiency | 7840 | (34.8%) |
| Yin-deficiency | 3670 | (16.3%) |
| Yang-deficiency | 1189 | (5.2%) |
| Phlegm-wetness | 2514 | (11.1%) |
| Wetness-heat | 2557 | (11.3%) |
| Qi-depression | 2059 | (9.1%) |
| Blood-stasis | 933 | (4.1%) |
| Special diathesis | 91 | (0.4%) |
| Gentleness | 1629 | (7.2%) |
| Total | 22482 | |

### 3.2 Experiment Setting

In this paper, we implement three kinds of deep learning methods: VGG-16, Inception-V3 and ResNet-50 models and modify their second-to-last layer to a fully-connected layer with 1024 units. All these networks are initialized with pre-trained model weights on ImageNet and train using Adam, where beta1 is 0.9 and beta2 is 0.99. We train these models using batch size 64 for 100 epochs without data augmentation. The initial learning rate is set to 0.0001. The output of the second-to-last layer of trained models are used to represent the tongue image and each image is mapped into a 1024-d feature vector.

In addition to deep convolutional neural network, we also utilize LBP to extract texture feature and Color-Moment to extract color feature. We split tongue image into $10 \times 10$ regions and utilize LBP on each region with 8 sampling points on a 1 radius circle. And then, the features are stitched together and reduced to 50 dimensions by using PCA. Color-Moment is used to calculate mean, variance and skewness in image to represent color distribution of tongue image. Finally, the feature of tongue image is represented as the combination of texture and color features.

In this paper, 5-fold cross-validation is applied to divide training set and testing set. Here, 15% of training set is used as validation set to adjust feature extraction model and hyper-parameter of Complexity Perception method. We implement three basis classifiers, Softmax, SVM and DecisionTree, to evaluate the effectiveness of our proposed method. According to the experimental experience, we set $(k=150, e=2, n=50)$ for convolutional neural network feature extraction method and $(k=5, e=40, n=50)$ for traditional feature extract method. Hyper-parameter $\theta$ that indicates the easy degree of training sample is determined by the validation set. The average classification accuracy of the 5 experiments is used to evaluate the performance of algorithm.

### 3.3 Experimental Result and Analysis

To demonstrate the effectiveness of the proposed method, we compare the accuracy of constitution classification with and without Complexity Perception method respectively. Table.2-Table.4 show the comparison results on datasets Tongue-100, Tongue-80 and Tongue-60 with optimal hyper-parameter respectively. Note that $Clf_B$ is a base classifier that trains on the whole training data, $Clf_E$ means easy classifier that only trains on easy training data and $Clf_D$ represents difficult classifier that only trains on difficult training data. In experiments, all testing samples are firstly divided into easy testing samples and the difficult testing samples using method described in the section 2.4.

In Table.2-Table.4, Easy_B is the classification accuracy of easy testing samples predicted by $Clf_B$, Easy_CP is the classification accuracy of the easy testing samples predicted by $Clf_E$, Diff_B is the classification accuracy of the difficult testing samples predicted by $Clf_B$ and Diff_CP is the classification accuracy of the difficult testing samples predicted by $Clf_D$. Basic means the result of whole testing data predicted by $Clf_B$, while CP is the result of Complexity Perception that combines easy classifier $Clf_E$ and difficult classifier $Clf_D$.

**Table 2** The average classification accuracy (percentage) of each method on Tongue-100

| Method | Comparison_E | | Comparison_D | | Comparison_B | |
|---|---|---|---|---|---|---|
| | Easy_B | Easy_CP | Diff_B | Diff_CP | Basic | CP |
| ResNet-50+Sofamax | 55.84 | **56.95** | 47.94 | **83.56** | 55.69 | **57.39** |
| ResNet-50+SVM | 56.75 | **57.09** | 34.24 | **86.30** | 56.38 | **57.56** |
| ResNet-50+DecisionTree | 53.01 | **55.34** | **55.33** | 50.70 | 52.83 | **54.83** |
| Inception-V3+Sofamax | 57.49 | **57.85** | 47.22 | **88.88** | 57.42 | **58.10** |
| Inception-V3+SVM | 57.64 | **57.89** | 48.52 | **62.72** | 57.30 | **58.08** |
| Inception-V3+DecisionTree | 54.44 | **56.24** | 86.84 | **89.47** | 54.66 | **56.52** |
| VGG-16+Sofamax | 58.88 | **59.42** | **62.14** | 59.32 | 59.03 | **59.41** |
| VGG-16+SVM | 59.44 | **59.82** | 77.77 | **88.88** | 59.55 | **59.99** |
| VGG-16+DecisionTree | 56.51 | **58.16** | 51.49 | **52.61** | 56.05 | **57.83** |
| LBP&Color-Moment+Sofamax | 51.30 | **53.91** | 35.53 | **35.58** | 37.82 | **38.25** |
| LBP&Color-Moment+SVM | 54.21 | **54.52** | 42.75 | **43.03** | 44.41 | **44.70** |
| LBP&Color-Moment+DecisionTree | 39.16 | **53.16** | 44.84 | 43.21 | 43.51 | **45.17** |

Table.2 depicts the experimental results on Tongue-100. It can be seen that Easy_CP is higher than Easy_B for all methods by 2.14% accuracy improvement on the average. On Comparison_D, Diff_CP is generally better than Diff_B, whose accuracy is increased by 12.47% on the average. Furthermore, it can be observed that our proposed method gives a better performance than basic methods under the following three classifiers. VGG-16 model obtains the best basic performance but the improvement of our proposed method is only 0.86%. By contrast, ResNet-50 obtains 1.62% improvement of classification accuracy on the average. LBP&Color-Moment not only has the lowest basic classification accuracy, but our proposed method can only increase the accuracy by 0.79%. It can be seen that, the proposed method performs best on DecisionTree. Compared with Softmax and SVM whose average accuracy is improved by 0.79% and 0.67%, Decision tree can improve the accuracy of 1.825% by using the proposed method.

The experimental results on dataset Tongue-80 are shown in Table.3. As can be seen, Easy_CP accuracy is higher than Easy_B on each methods and improves 3.73% on the average. Improved accuracy can be observed on most methods from Comparision_D. Note that, our proposed method is higher than basic method by 1.05% on the average on Tongue-80. From Table.3, it can be seen that the highest classification accuracy improvement of Tongue-80 is 1.91% by using ResNet-50+DecisionTree. Our proposed method can obtain better classification performance improvement than Softmax and SVM by using DecisionTree. Compared with neural network, LBP&Color-Moment is only 0.65% higher than the base method, while ResNet-50、Inception-V3 and VGG-16 obtain 1.07%、1.12% and 1.36% higher performance on the average respectively.

**Table 3** The average classification accuracy (percentage) of each method on Tongue-80

| Method | Comparison_E | | Comparison_D | | Comparison_B | |
|---|---|---|---|---|---|---|
| | Easy_B | Easy_CP | Diff_B | Diff_CP | Basic | CP |
| ResNet-50+Sofamax | 52.44 | **52.90** | 59.7 | **64.17** | 52.73 | **53.32** |
| ResNet-50+SVM | 52.55 | **53.14** | 80.64 | **96.77** | 52.79 | **53.51** |
| ResNet-50+DecisionTree | 48.64 | **50.01** | 93.75 | **100.0** | 48.32 | **50.23** |
| Inception-V3+Sofamax | 52.84 | **53.60** | 53.33 | **86.66** | 52.8 | **53.74** |
| Inception-V3+SVM | 53.73 | **54.02** | 45.59 | **50.25** | 53.29 | **53.82** |
| Inception-V3+DecisionTree | 49.84 | **51.81** | 57.84 | **61.76** | 50.21 | **52.10** |
| VGG-16+Sofamax | 52.23 | **54.16** | 48.68 | **50.00** | 52.29 | **54.07** |
| VGG-16+SVM | 53.61 | **54.80** | **49.06** | 46.87 | 53.21 | **54.10** |
| VGG-16+DecisionTree | 50.46 | **52.17** | **53.59** | 47.71 | 50.57 | **51.98** |
| LBP&Color-Moment+Sofamax | 50.00 | **66.67** | 37.17 | **37.22** | 37.19 | **37.27** |
| LBP&Color-Moment+SVM | 49.22 | **49.22** | 40.78 | **41.11** | 42.28 | **42.56** |
| LBP&Color-Moment+DecisionTree | 34.50 | **52.40** | 38.18 | **38.98** | 38.24 | **39.83** |

**Table 4** The average classification accuracy (percentage) of each method on Tongue-60

| Method | Comparison_E | | Comparison_D | | Comparison_B | |
|---|---|---|---|---|---|---|
| | Easy_B | Easy_CP | Diff_B | Diff_CP | Basic | CP |
| ResNet-50+Sofamax | 47.92 | **49.93** | 41.98 | **42.36** | 47.37 | **49.20** |
| ResNet-50+SVM | 47.99 | **49.56** | 42.48 | **44.44** | 47.68 | **49.27** |
| ResNet-50+DecisionTree | 43.96 | **45.90** | 50.45 | **52.29** | 43.91 | **46.15** |
| Inception-V3+Sofamax | 48.76 | **49.64** | 30.35 | **33.92** | 48.39 | **49.31** |
| Inception-V3+SVM | 49.04 | **49.88** | 36.36 | **38.96** | 48.68 | **49.57** |
| Inception-V3+DecisionTree | 46.26 | **47.90** | 35.29 | **41.17** | 46.33 | **47.86** |
| VGG-16+Sofamax | 48.84 | **50.04** | **38.74** | 37.26 | 47.81 | **48.75** |
| VGG-16+SVM | 49.35 | **50.90** | 38.87 | **40.21** | 47.90 | **49.42** |
| VGG-16+DecisionTree | 46.59 | **49.18** | 37.80 | **38.87** | 45.16 | **47.75** |
| LBP&Color-Moment+Sofamax | **44.93** | 40.58 | 36.18 | **36.44** | 36.40 | **36.55** |
| LBP&Color-Moment+SVM | 45.71 | **51.43** | 41.02 | **51.05** | 41.07 | **41.19** |
| LBP&Color-Moment+DecisionTree | 32.95 | **45.36** | **35.43** | 30.07 | 34.73 | **35.51** |

Table.4 shows the experimental results on dataset Tongue-60. It can be seen from Table.4 that Easy_CP perform better than Easy_B on most methods. Similar conclusions can also be obtained from the Comparison_D. Our proposed method can bring an average accuracy increment of 1.25% on Tongue-60. As shown in Table.4, VGG-16+DecisionTree can achieve highest improvement of classification accuracy. With our proposed method, CP+DecisionTree is 1.78% higher than the basic DecisionTree on Tongue-60. From the result of Comparison_D and Comparison_E, Diff_CP is higher than Diff_B by 1.84% and Easy_CP is higher than Easy_B by 2.33% on the average. Our proposed method can obtain 1.56% accuracy improvement on neural network feature extraction framework, while traditional feature extraction method can only achieve 0.35% accuracy improvement. However, methods based on deep neural network are often better than traditional feature extraction methods.

### 3.4 Effective of hyper-parameter

Fig.4-Fig.7 depict the performance of our proposed method on three datasets with the change of easy degree threshold $\theta$ in the range of 0.05-0.95.

Fig.4 (a), (b) and (c), respectively, depicts the performance curves with the increment of $\theta$ on Tongue-100, Tongue-80 and Tongue-60 by using ResNet-50. It can be observed that, although the accuracy of our proposed method is undulate, the performance of our proposed method is higher than the baseline on the most values of $\theta$, which illustrates that our proposed method can effectively improve the performance of classification. Compared to Softmax and SVM classifiers, DecisionTree fluctuates significantly, but has a greater increase in accuracy. As the size of the dataset shrinks, the classification accuracy decreases, but the method we propose still remains effective. As shown in Fig.4, under the same dataset, the threshold $\theta$ for the three classifiers to obtain optimal results are different. Moreover, for the same classifier, the optimal threshold $\theta$ on three datasets are different.

Fig.5 shows the performance of Complexity Perception method with the increment of $\theta$ by using Inception-V3

feature extract framework. Our proposed method performs better than the baseline on the most values of threshold $\theta$. It can be seen that, DecisionTree obtains the highest classification accuracy improvement on the average.

We conduct our experiments by VGG-16 feature extraction framework on three datasets and the results are shown in Fig.6. Our proposed method can achieve higher classification accuracy than the baseline method for the most values of $\theta$. As shown in Fig.6, the highest accuracy improvement on Tongue-80 is achieved using Softmax classifier, while DecisionTree performs best on another two datasets.

Fig.7 shows the classification accuracy versus the parameter of easy degree threshold $\theta$ by using LBP&Color-Moment. As shown in Fig.7, our proposed method only performs better than the base method with a few easy degree thresholds. It is hard to find an appropriate parameter of $\theta$ for LBP&Color-Moment feature extraction framework. It can be seen that the classification accuracy of DecisionTree decreases obviously with the size of dataset declines.

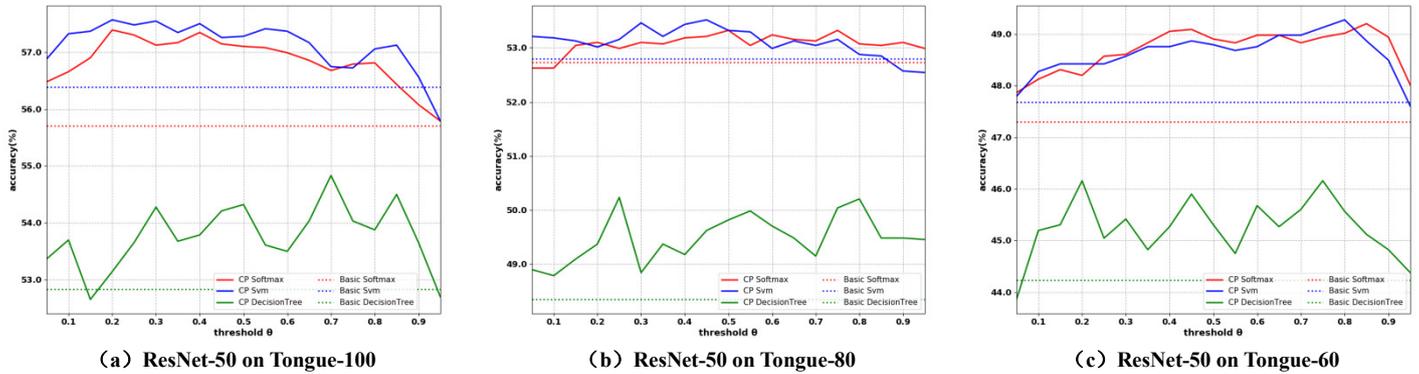

(a) ResNet-50 on Tongue-100　　(b) ResNet-50 on Tongue-80　　(c) ResNet-50 on Tongue-60

**Fig. 4.** Testing-accuracy for ResNet-50 feature extract framework on 3 datasets

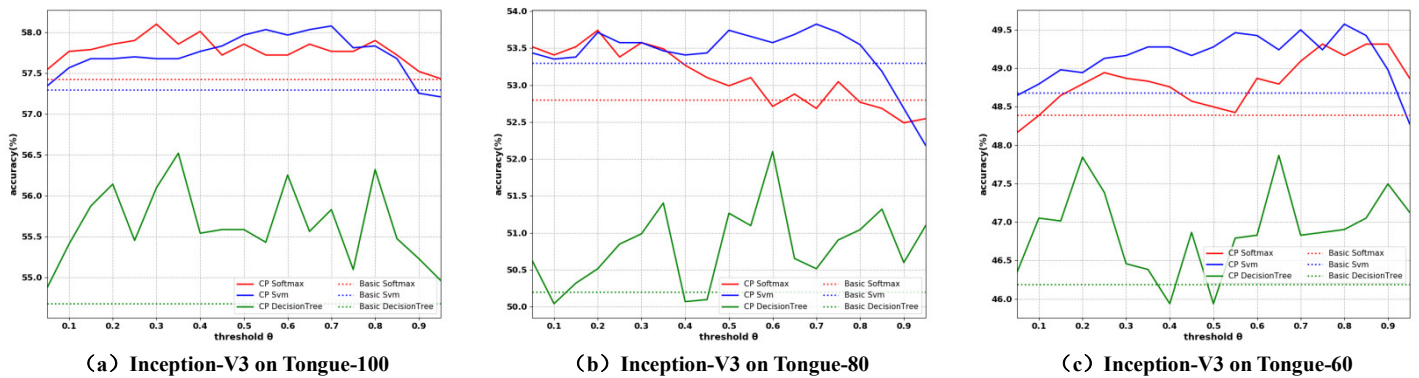

(a) Inception-V3 on Tongue-100　　(b) Inception-V3 on Tongue-80　　(c) Inception-V3 on Tongue-60

**Fig. 5.** Testing-accuracy for Inception-V3 feature extract framework on 3 dataset

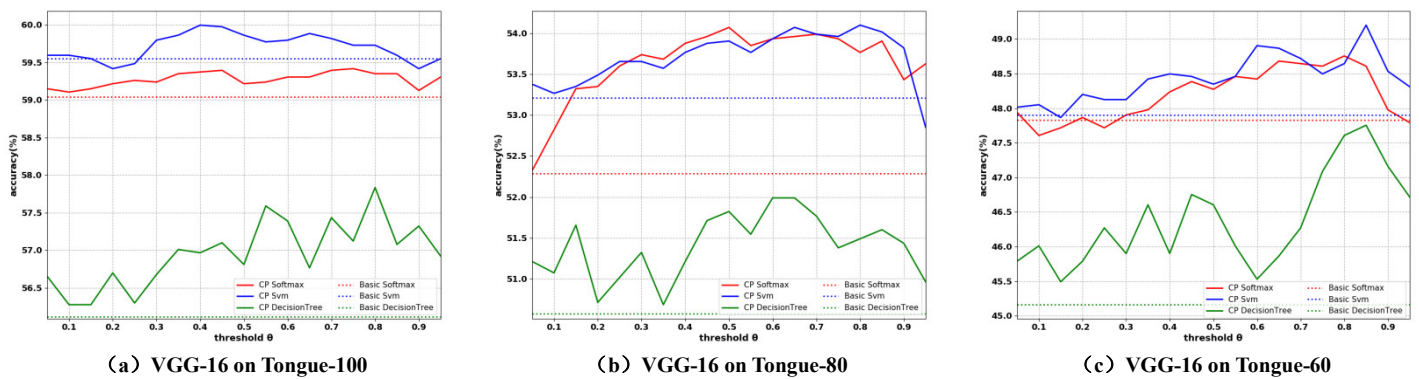

(a) VGG-16 on Tongue-100　　(b) VGG-16 on Tongue-80　　(c) VGG-16 on Tongue-60

**Fig. 6.** Testing-accuracy for VGG-16 feature extract framework on 3 datasets

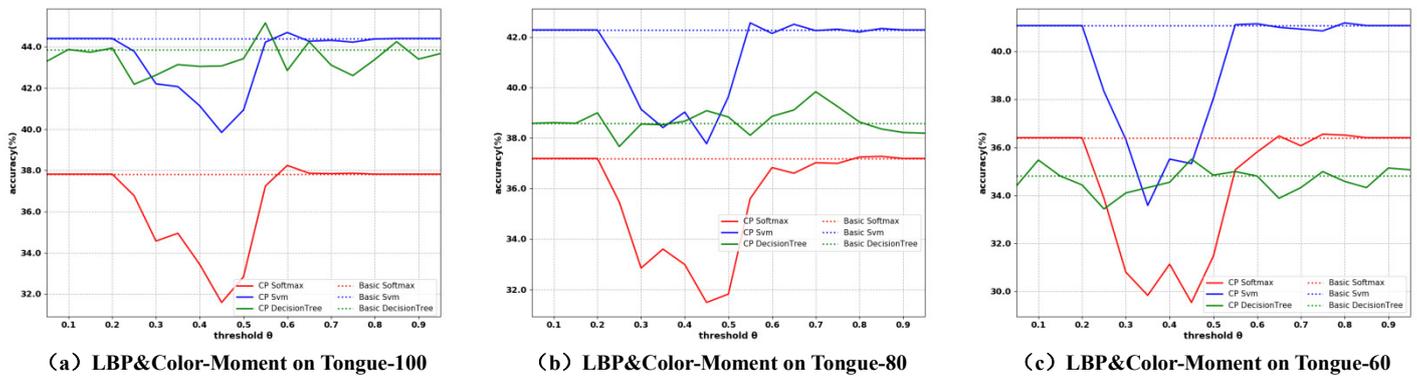

**Fig. 7.** Testing-accuracy for LBP&Color-Moment feature extract framework on 3 datasets

## 4. Discussion

We evaluate the performance of our proposed method on three tongue datasets by using four feature extract framework, respectively. From the experimental results, we can see that our method is effective on all three datasets, and the method based on deep convolutional neural network is generally better than the traditional machine learning methods. From the results of Table.2-Table.4, we find that Easy_CP is higher than Easy_B by 2.73% on the average. Diff_CP is higher than Diff_B by 6.52%. These indicate that separating data into two subsets according to their complexity can effectively improve classification accuracy. From the results of Comparison_D, the difference between Diff_CP and Diff_B on Tongue-100, Tongue-80 and Tongue-60 is 12.47%, 5.26% and 1.84%, respectively. It can be seen that the size of dataset affects the ability to distinguish easy samples or difficult samples. Note that, the proposed method performs best on DecisionTree classifier. From the results of Fig.4-Fig.6, our proposed method combined with the features extracted by deep convolutional neural network can effectively improve the classification accuracy. However, Fig.7 depicts that the proposed method has no obvious effect on the features extracted by LBP&Color-Moment. Therefore, our proposed method has requirement for the method of feature extraction. Indeed, compared to traditional feature extraction methods, our proposed method is more suitable for neural network feature extraction method. The results of basic Softmax classifiers in the experiments are basically equivalent to the performance of convolutional neural network and the proposed method can improve the performance again. Therefore, we propose a method that combines Complexity Perception with deep learning to improve the classification performance of tongue images constitution recognition.

Although the experiments are evaluated on Chinese patients' tongue images datasets, we believe that the proposed method can be easily extended to other datasets and application scenarios.

## 5. Conclusions

In this work, we design a tongue images constitution recognition framework, including images acquisition, detection, calibration and classification to achieve accurate, fast and efficient constitution recognition. We introduce Complexity Perception method that divides data into two subsets according to their complexity at individual level and then processes them separately. The effectiveness of the proposed Complexity Perception is evaluated using three tongue image datasets and four feature extraction methods. The experimental results demonstrate that, compared to the baseline, our proposed method can effectively improve the performance of tongue images constitution recognition. To the best of our knowledge, this is the first attempt in applying computer-aid tongue diagnosis to recognize constitution types. The experimental results show that the best classification accuracy our method obtained is 59.99%, which can be acceptable by many doctors in hospital. Our proposed framework and method can provide important support for doctor's diagnosis.

In the future, we plan to further explore the performance of our proposed method, e.g., for different complexity of the samples, using different feature extraction methods or using different pre-process methods.

## Conflict of Interest statement

The authors declare that there is no conflict of interests regarding the publication of this article.

## Acknowledgements

This work was supported by a China National Science Foundation under Grants (81274110, 60973083, 61273363), Science and Technology Planning Projects of Guangdong Province (2014A010103009, 2015A020217002), and Guangzhou Science and Technology Planning Project (201504291154480)